# Machine learning on small size samples: A synthetic knowledge synthesis

Peter Kokol, Marko Kokol*, Sašo Zagoranski*

*Faculty of Electrical Engineering and Computer Science, University of Maribor, Maribor, Slovenia*

**Semantika, Maribor, Slovenia*

## ITRODUCTION

Periods of scientific knowledge doubling have become significantly becoming shorter and shorter (R. Buckminster Fuller, 1983; Bornmann, Anegón and Leydesdorff, 2010; McDeavitt, 2014). This phenomena, combined with information explosion, fast cycles of technological innovations, Open Access and Open Science movements, and new Web/Internet based methods of scholarly communication have immensely increased the complexity and effort needed to synthesis scientific evidence and knowledge. However, above phenomena also resulted in the growing availability of research literature in a digital, machine-readable format.

To solve the emerging complexity of knowledge synthesis and the simultaneous new possibilities offered by digital presentation of scientific literature, Blažun et all (Blažun Vošner *et al.*, 2017) and Kokol et al (Kokol *et al.*, 2020; Kokol, Zagoranski and Kokol, 2020) developed a novel synthetics knowledge synthesis methodology based on the triangulation of (1) distant reading (Moretti, 2013), an approach for understanding the canons of literature not by close (manual) reading, but by using computer based technologies, like text mining and machine learning, (2) bibliometric mapping (Noyons, 2001) and (3) content analysis (Kyngäs, Mikkonen and Kääriäinen, 2020; Lindgren, Lundman and Graneheim, 2020). Such triangulation of technologies enables one to combine quantitative and qualitative knowledge synthesis in the manner to extend classical bibliometric analysis of publication metadata with the machine learning supported understanding of patterns, structure and content of publications (Atanassova, Bertin and Mayr, 2019).

One of the increasingly important technologies dealing with the growing complexity of the digitalisation of almost all human activities is the Artificial intelligence, more precisely machine learning (Krittanawong, 2018; Grant *et al.*, 2020). Despite the fact, that we live in a "Big data" world where almost "everything" is digitally stored, there are many real world situation, where researchers are faced with small data samples. Hence, it is quite common, that the database is limited for example by the number of subjects (i.e. patients with rare disseises), the sample is small comparing to number of features like in genetics or biomarkers detection, sampling, there is a lot of noisy or missing data or measurements are extremely expensive or data iimbalanced meaning that the size of one class in a data set has very few objects (Thomas *et al.*, 2019; Vabalas *et al.*, 2019).

Using machine learning on small size datasets pose a problem, because, in general, the "power"of machine learning in recognising patterns is proportional to the size of the dataset, the smaller the dataset, less powerful and less accurate are the machine learning algorithms. Despite the commonality of the above problem and various approaches to solve it we didn't found any holistic studies concerned with this important area of the machine learning. To fill this gap we used synthetic knowledge synthesis presented above to aggregate current evidence relating to the »small data set problem« in machine learning. In that manner we can overcome the problem of isolated findings which might be incomplete or might lack possible overlap with other possible solutions. In our analysis, we aimed to extract, synthetise and multidimensionally structure the evidence of as possible complete corpus of scholarship on small data sets. Additionally, the study seeks to identify gaps which may require further research.

Thus the outputs of the study might help researchers and practitioners to understand the broader aspects of small size sample` research and its translation to practice. On the other hand, it can help a novice or data science professional without specific knowledge on small size samples to develop a perspective on the most important research themes, solutions and relations between them.

## METHODOLOGY

Following research question research questions was set

*What is the small data problem and how it is solved?*

Synthetic knowledge synthesis was performed following the steps below:

1. Harvest the research publications concerning small data sets in machine learning to represent the content to analyse.
2. Condense and code the content using text mining.
3. Analyse the codes using bibliometric mapping and induce the small data set research cluster landscape.
4. Analyse the connections between the codes in individual clusters and map them into sub-categories.
5. Analyse sub-categories to label cluster with themes.
6. Analyse sub-categories to identify research dimensions
7. Cross-tabulate themes and research dimension and identify concepts

The research publications were harvested from the Scopus database, using the advance search using the command *TITLE-ABS-KEY(("small database" or "small dataset" or "small sample" ) and "machine learning" and not ("large database" or "large dataset" or "large sample" ))* The search was performed on 7th of January, 2021. Following metadata were exported as a CSV formatted corpus file for each publication: Authors, Authors affiliations, Publication Title, Year of publication, Source Title, Abstract and Author Keywords.

Bibliometric mapping and text mining were performed using the VOSViewer software (Leiden University, Netherlands). VOSViewer uses text mining to recognize publication terms and then employs the mapping technique called Visualisation of Similarities (VoS) to create bibliometric maps or landscapes (van Eck and Waltman, 2014). Landscapes are displayed in various ways to present different aspects of the research publications content. In this study the content analysis was performed on the author keywords cluster landscape, due to the fact that previous research showed that authors keywords most concisely present the content authors would like to communicate to the research community (Železnik, Blažun Vošner and Kokol, 2017). In this type of landscape the VOSviewer merges author keywords which are closely associated into clusters, denoted by the same cluster colours (van Eck and Waltman, 2010). Using a customized Thesaurus file, we excluded the common terms like study, significance, experiment and eliminated geographical names and time stamps from the analysis.

## RESULTS AND DISCUSSION

The search resulted in 1254 publications written by 3833 authors. Among them were 687 articles, 500 conference papers, 33 review papers, 17 book chapters and 17 other types of publications. Most productive countries among 78 were China (n=439), United States (n=297), United Kingdom (n=91), Germany (n=63), India (n=52), Canada (n=51) and Spain (n=44), Australia (n=42), Japan (n=34), South Korea (n=30) and France (n=30). Most productive institutions are located in China and United States. Among 2857 institution, China Academy of Sciences is the most productive institution (n=52), followed by Ministry of Education China (n=28), Harvard Medical School, USA (n=14), Harbin Institute of Technology, China (n=14), National Cheng Kung University, China (13), Shanghai University, China (n=13), Shanghai Jiao Tong University, China (n=12), Georgia Institute of Technology, USA (n=12) and Northwestern Polytechnical University, China (n=11). The first non China/USA institution is the University of Granada (n=8) on 19th rank. Majority of the productive countries are member of G9 countries, with strong economies and research infrastructure.

Most prolific source titles (journals, proceedings, books) are the Lecture Notes In Computer Science Including Subseries Lecture Notes In Artificial Intelligence And Lecture Notes In Bioinformatics (n=64), IEEE Access (n=17), Neurocomputing (n=17), Plos One (n=17), Advances in Intelligent Systems and Computing (n=14), Expert System with application (n=14), ACM International Conference Proceedings Series (n=13) and Proceedings Of SPIE The International Society For Optical Engineering (n=13). The more productive source titles are mainly from the computer science research area and are ranked in the first half of Scimago Journal ranking (Scimago, Elsevier, netherlands). In average their h-indexs is between 120 an 380.

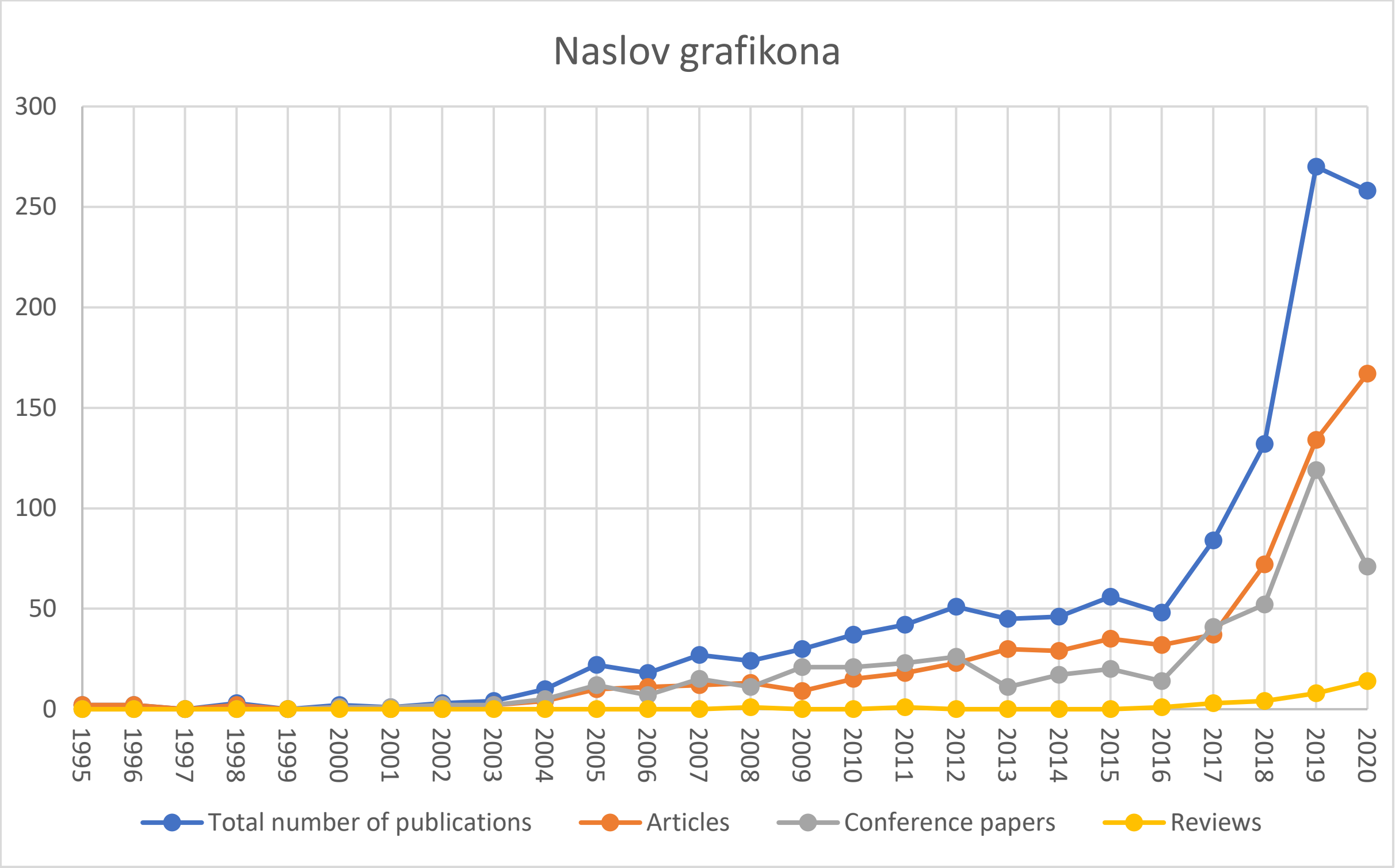

Figure 1. The research literature production dynamics

First two papers concerning the use of machine learning on small datasets indexed in Scopus were published in 1995 (Forsström *et al.*, 1995; Reich and Travitzky, 1995). After that the publications were rare till 2002, trend starting to rise linearly in 2003, and exponentially in 2016. In the beginning till the year 2000 all publications were published in journals, after that conference papers started to emerge. Uninterrupted production of review papers started much later in 2016. According to the ratio between articles, reviews and conference paper and the exponential trend in production for the first two we asses that machine learning on small datasets in the third stage of Schneider scientific discipline evolution model (Shneider, 2009). That means that the terminology and methodologies are already highly developed, and that domain specific original knowledge generation is heading toward optimal research productivity.

On the other hand Pestana et al (Pestana, Sánchez and Moutinho, 2019) characterised the scientific discipline maturity as a set of extensive repeated connections between researchers that collaborate on the publication of papers on the same or related topics over time. In our study we analysed those connection with the co-authors networks induced by VOSviewer. The co-author network showed that among countries with the productivity of 10 or more papers (n=27) presented in Figure 2. an elaborate international co-authorship network emerged, confirming that maturity is already reached a high level. The most intensive co-operation is  between United states and China, and the "newest" countries joining the network are Russian federation, Soth Korea and Brasil.

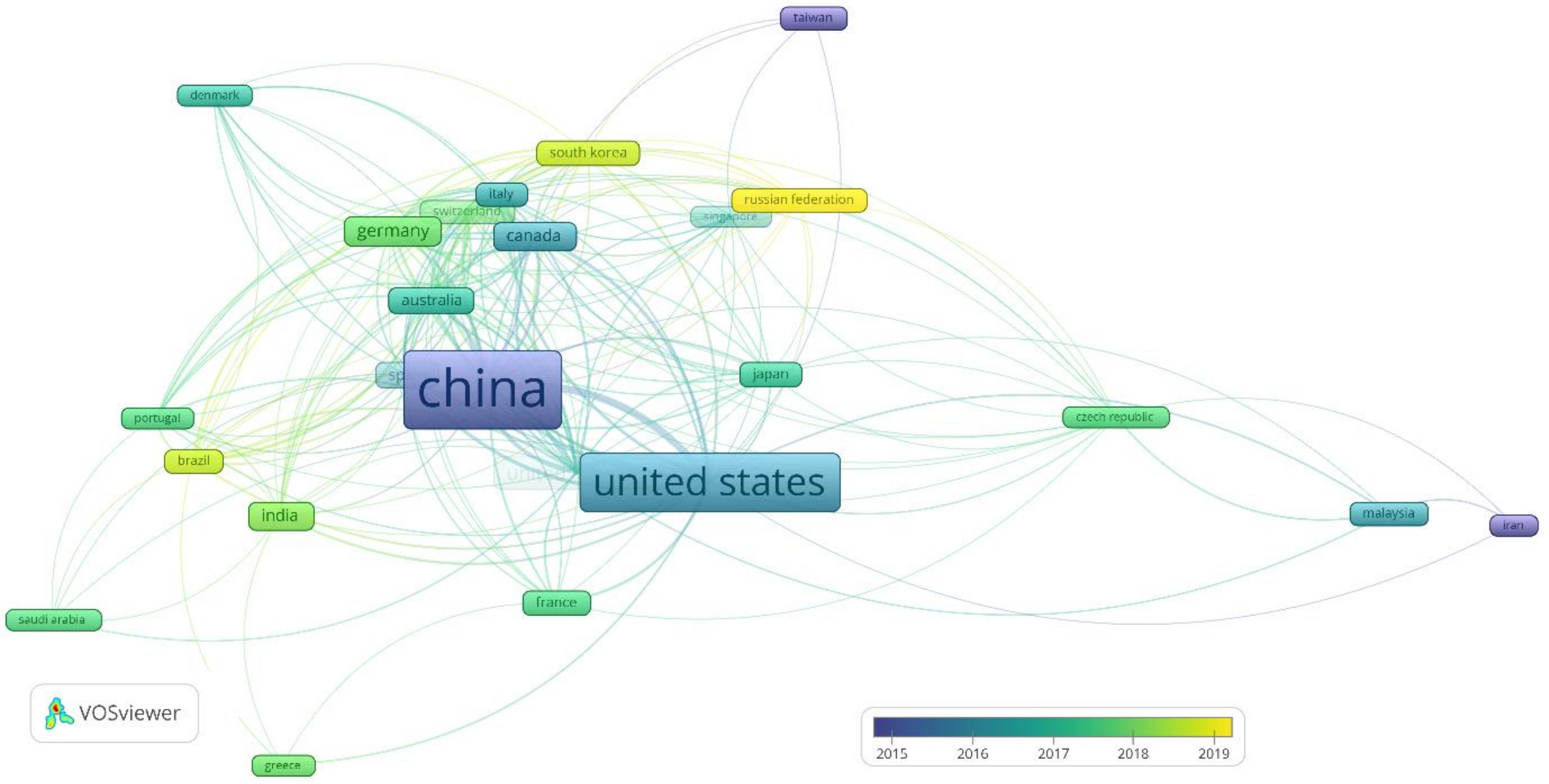

Figure 2. The co-authorship network

The content analysis of keyword cluster landscape shown in Figure 3. resulted in codes, SUB-categories and themes presented in Table 1. The content analysis revealed four prevailing themes. The largest two themes are related to dimension reduction in complex big data analysis and data augmentation techniques in deep learning.

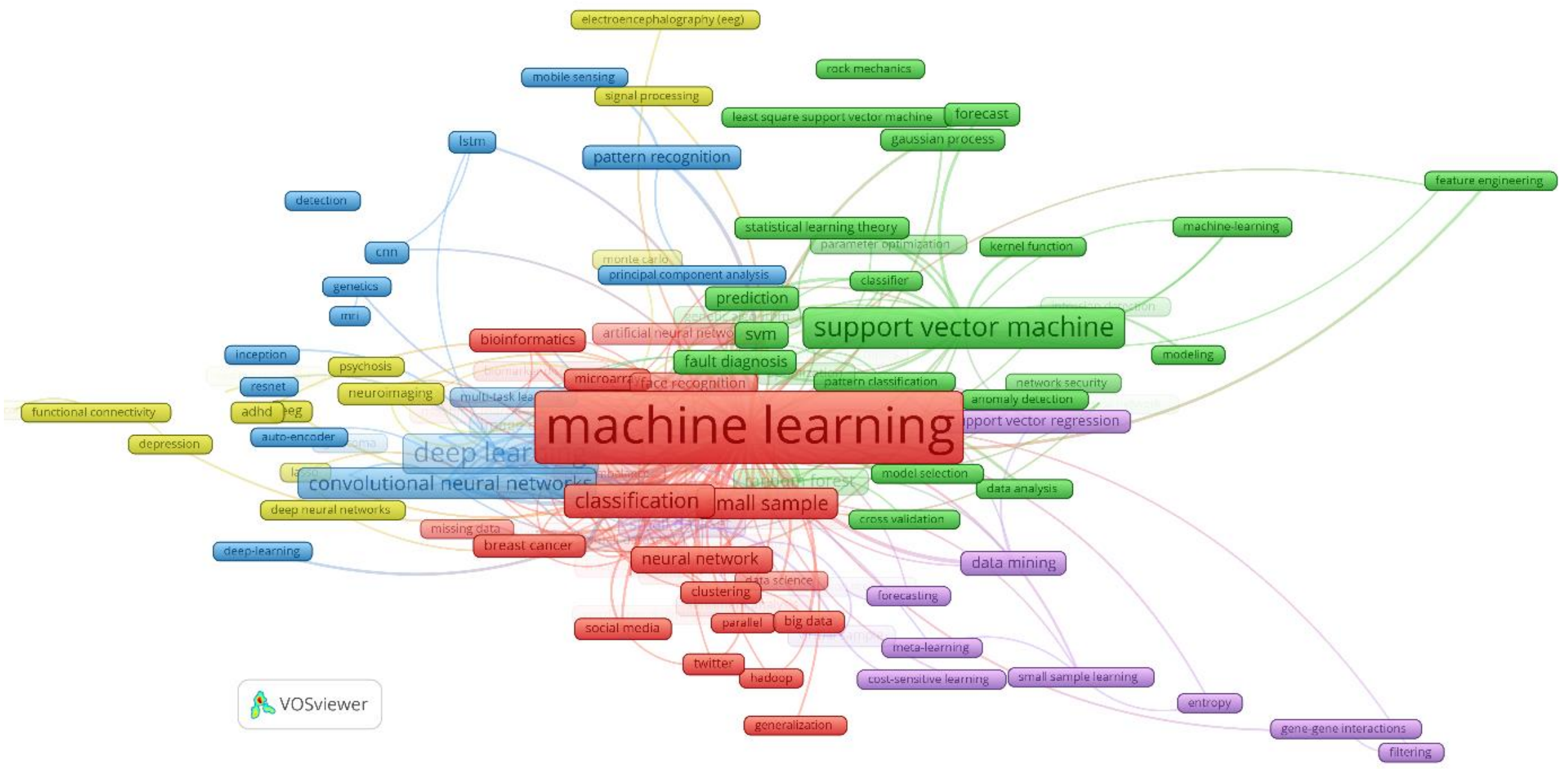

Figure 3. The author keyword cluster landscape

Table 1: Small size sample research themes (numbers in parenthesis present the number of papers in which an author keyword occurred)

| Theme | Colour | More frequent codes | Prevailing sub-categories |
|---|---|---|---|
| Dimension reduction in complex big data analysis | Red | Machine learning (339), Classification (51), Feature selection (50), Artificial neural network (28), Neural network (19), Natural language processing (16), Ensemble learning (14), Bioinformatics (13), Decision tree (12), Breast cancer (11), Machine learning algorithms (10), Computer | Machine learning algorithms in bioinformatics, Classification on small samples with feature selection, Ensemble learning for solving under-sampling in personalised medicine, Solving missing data in breast cancer using |
| | | vision (10), Microarray (9), Small datasets (9), Text classification (9), Big data (9), Dimensionality reduction (9), Social media (7), | deep neural networks, Dimensionality reduction. with feature extraction in cancer classification, Natural language processing in social media and text analysis |
| Data augmentation techniques in deep learning for pattern recognition and classification | Blue and yellow | Deep learning (89), Convolutional neural networks (45), Transfer learning (38), Pattern recognition (14), Image classification (14), data augmentation (12), Generative adversarial networks (8), LSTM (6), Resnet – Residual neural network (n=4) | Deep and transfer learning using in image classification using data augmentation and synthetic data, complex neural networks, Pattern recognition in mobile sensing using Principal component analysis and LTSM, Resnet and Desnet based deep learning using auto encoder on case of glaucoma, |
| Support vector machines, Random forest and Genetic algorithms in statistical learning | Green | Support vector machines (109), Random forest (24), Prediction (18), Fault diagnosing (18), Forecast (10), Genetic algorithms (8), Regression (8), Gaussian process (7), Optimisation (7), Time series (6), Statistical learning theory (6) | Support vector machines and random forests in prediction, forecasting and fault diagnosis, Genetic algorithms in optimisation and time series analysis |
| Data mining on small datasets with support | Viollet | Small datasets (2), Data mining (15), Support vector regression (11), Virtual sample (7) | Small datasets augmentation with virtual samples, Data mining with support vector regression |
| vectors regression | | | |

The higher abstraction analysis of the Table 1. revealed four categories, which we call Research dimensions. Four identified dimensions are: Small data set problem, Machine learning algorithms, Small-data pre-processing technique and Application Area. Cross tabulation of themes and research dimension resulted in a taxonomy presented in Table 2. The entries in the table presents the most popular concepts in each taxonomical entity.

The most frequently reported difficulties causing the small data problem are the small size of the dataset, high/low dimensionality of datasets and unbalanced data. Small size datasets can cause problems when machine learning is applied in material sciences (Zhang and Ling, 2018), engineering (Babič *et al.*, 2014; Feng, Zhou and Dong, 2019) and various omics fields (Ko, Choi and Ahn, 2021) due to the high cost of sampling; differentiating between autistic and non-autistic patients (Vabalas *et al.*, 2019) and diagnosing rare diseases (Spiga *et al.*, 2020) due to the small number of available patients, or unavailability of other subjects for example PhD students in prediction of their grades (Abu Zohair, 2019); and new pandemic prediction due fact that samples are scare when pandemic occurs and there is lack of medical knowledge (Fong *et al.*, 2020). Similar reasons can also lead to too high or low dimensionality of the datasets. The first case can occur even if the sample size is not small, however the ratio between number of features and the sample size is large, like for example in particle physics (Komiske *et al.*, 2018) or bioinformatics (Zhang *et al.*, 2021). The second case can occur when the number of instances is not problematic, however the number of features is very small, like for example in characterisation of high-entropy alloys (Dai *et al.*, 2020). Unbalanced data present a long standing problem in machine learning and still remains a challenge in various applications like face recognition, credit scoring, fault diagnosing and anomaly detection where mayor class has much more instances than one or more of remaining classes (He *et al.*, 2020; Liang *et al.*, 2020; Marceau *et al.*, 2020).

The most used machine learning algorithms used on small datasets are support vector machines (Alvarsson *et al.*, 2016; Cao *et al.*, 2017; Razzak *et al.*, 2020), decision trees/forests (Zhang *et al.*, 2017; Shaikhina *et al.*, 2019), convolutional neural networks (Liu and Deng, 2016; Yamashita *et al.*, 2018; Guo *et al.*, 2020) and transfer learning (Hall *et al.*, 2020; Tits, El Haddad and Dutoit, 2020).

Most frequently employed data pre-processing techniques to overcame the small size problem are liner and nonlinear Principal component analysis (Jalali, Mallipeddi and Lee, 2017; Feng, Zhou and Dong, 2019; Athanasopoulou, Papacharalampopoulos and Stavropoulos, 2020), Discriminant analysis (Abu Zohair, 2019; Li *et al.*, 2019; Silitonga *et al.*, 2021), Data augmentation(Han, Liu and Fan, 2018; Hagos and Kant, 2019; Fong *et al.*, 2020), Virtual sample (Gong *et al.*, 2017; MacAllister, Kohl and Winer, 2020; Zhu *et al.*, 2020), Feature extraction (Kumar *et al.*, 2018; Dai *et al.*, 2020) and Auto-encoder (Feng, Zhou and Dong, 2019; Pei *et al.*, 2021).

Most affected areas are Bioinformatics (Giansanti *et al.*, 2019; Khatun *et al.*, 2020; Ko, Choi and Ahn, 2021), image classification and analysis (Han, Liu and Fan, 2018; Hagos and Kant, 2019; Yadav and Jadhav, 2019), fault diagnosing (Saufi *et al.*, 2020; Wen, Li and Gao, 2020), forecasting and prediction (Croda, Romero and Morales, 2019; Fong *et al.*, 2020), social media analysis (Khan *et al.*, 2019; Renault, 2020) and health care (Rajpurkar *et al.*, 2020; Subirana *et al.*, 2020).

Table 2: Taxonomy of Themes and Research dimensions categories (numbers in parenthesis present the number of publications)

| Theme | Small data size problem | Machine learning algorithms | Small – data pre-processing technique | Application area |
|---|---|---|---|---|
| Dimension reduction in complex big data analysis | Missing data (3), Unbalanced data (8), Under sampling (3) | Bagging (9), Bayesian networks (7), Boosting (6), Decision trees and forests (43), deep neural network (12), Ensemble learning (21), manifold learning (5), neural network (30), naïve Bayes (2), semi supervised learning (17), supervised learning | Dimensionality reduction (9), Feature extraction (12), engineering (4), fusion (2), Lasso (6), Linear and multiple discriminant analysis (13), PCA (16), | Image analysis (18), genetics and bioinformatics (27), personalised and precision medicine (6) neuroimaging (7), mental health (25) |
| | | (19), unsupervised learning (4), FMRI (7) | autoencoder (10) | |
| Data augmentation for deep learning in pattern recognition and classification | | Convolutional neural network (42), Dense CNN (3), Generative adversarial networks (8), LSTM (6), Resnet (4), transfer learning (40), semi-supervised learning (17) | Data augmentation (12), Domain adaption (5), data fusion (3), synthetic data (3), autoencoder (10) | Image classification (22), genetics (3), mobile sensing (10), predictive modelling (6), semantic segmentation (3) |
| Statistical based machine learning | Small sample (98) | Support vector machines (89). Least square SVM (10), genetic algorithms, random forest (24), Particle swarm (27), support vector regression (11) | Feature engineering (3), meta-learning (5), Monte Carlo (3), Small sample learning (5), virtual sample (11) | Forecasting (13), Fault diagnosing (18), regression (19), prediction (26), pattern recognition (17), rock mechanics (5), security 7), computer vision (10), data mining (15) |
| Data mining on small datasets | Big data (9) | | | Text and social media analysis (21), Precision medicine (3), |
| | | | | Natural language processing (16), Sentiment analysis (9), clustering (8) |

*Strengths and limitations*

The main strength of the study is that it is the first bibliometrics and content analysis of the small dataset research. One of the limitation is that the analysis was limited to publications indexed in Scopus only, however Scopus is indexing the largest and most complete set of information titles, thus assuring the integrity of the data source. Additionally, the analysis alsp included qualitative components, which might bias the results of our study.

**Conclusion**

Our bibliometric study showed the positive trend in the number of research publications concerning the use of small dataset and substantial growth of research community dealing with the small dataset problem, indicating that the research field is moving toward the higher maturity levels. Despite notable international cooperation, regional concentration of research literature production in economically more developed countries was observed.

The results of study present a multi-dimensional facet and science landscape of the small datasets problem which can help machine learning researchers and practitioners to improve their understanding of the area and can catalyse further knowledge development. On the other hand, it can inform novice researchers, interested readers or research mangers and evaluators without specific knowledge and help them  to develop a perspective on the most important small dataset research dimensions. Finally, the study output can serve as a guide to further research.